\documentclass[journal]{IEEEtran}
\usepackage[T1]{fontenc}
\usepackage{graphicx}
\usepackage{times}
\usepackage{helvet}
\usepackage{courier}
\usepackage{amsmath}
\usepackage{algorithm}
\usepackage{algorithmic}
\usepackage{csquotes} 
\usepackage{color}
\usepackage{paralist}
\usepackage{amssymb}
\usepackage{indentfirst}
\usepackage{subfigure}
\usepackage{float}
\usepackage{multirow}
\usepackage{cite}
\usepackage{mathrsfs}

\usepackage{amsfonts}
\usepackage{todonotes}
\usepackage{pgfplots}
\usepackage{epstopdf}
\usepackage{epsfig}
\pgfplotsset{compat=newest}
\definecolor{forestgreen}{RGB}{0,139,69}

\usepackage{xcolor}
\definecolor{citecolor}{HTML}{0071bc}
\usepackage[colorlinks, linkcolor=red,  anchorcolor=blue, citecolor=citecolor]{hyperref} 

\definecolor{SeaGreen4}{RGB}{0,205,102} 
\definecolor{SlateBlue}{RGB}{106,90,205} 
\definecolor{DarkRed}{RGB}{178,34,34} 
	
\usepackage[switch]{lineno}

\usepackage{textcomp,booktabs}
\usepackage{pifont}
\newcommand{\eg}{e.g.}

\usepackage{makecell}

\usepackage{colortbl}
\definecolor{mygray}{gray}{.9}
\definecolor{mypink}{rgb}{.99,.91,.95}
\definecolor{mycyan}{cmyk}{.3,0,0,0}

\begin{document}

\title{
     T2I-VeRW: Part-level Fine-grained Perception for Text-to-Image Vehicle Retrieval 
} 

\author{Xiao Wang, \emph{Member, IEEE}, Ziwen Wang, Weizhe Kong, Wentao Wu, Yuehang Li, \\ 
        Aihua Zheng, Chenglong Li*, \emph{Senior Member, IEEE}, Jin Tang     

\thanks{Ziwen Wang, Yuehang Li, Xiao Wang, and Jin Tang are with the School of Computer Science and Technology, Anhui University, Hefei 230601, China (email: \{e24201001, e23201112\}@stu.ahu.edu.cn, \{xiaowang, tangjin\}@ahu.edu.cn)} 

\thanks{Weizhe Kong, Wentao Wu, Aihua Zheng, Chenglong Li are with the School of Artificial Intelligence, Anhui University, Hefei 230601, China. (email: weizhekong99@gmail.com, wa22201027@stu.ahu.edu.cn, \{ahzheng214, lcl1314\}@foxmail.com)} 

\thanks{* Corresponding Author: Chenglong Li (lcl1314@foxmail.com)} 
}

\markboth{IEEE Transactions on Intelligent Transportation Systems, 2026}
{Wang \MakeLowercase{\textit{et al.}}: Cross-modal Vehicle Re-Identification with Part-Level Fine-grained Perception}

\maketitle

\begin{abstract}
Vehicle Re-identification (Re-ID) aims to retrieve the most similar image to a given query from images captured by non-overlapping cameras. Extending vehicle Re-ID from image-only queries to text-based queries enables retrieval in real-world scenarios where only a witness description of the target vehicle is available. In this paper, we propose PFCVR, a Part-level Fine-grained Cross-modal Vehicle Retrieval model for text-to-image vehicle re-identification. PFCVR constructs locally paired images and texts at the part level and introduces learnable part-query tokens that aggregate both part-specific and full-sentence context before aligning with visual part features. On top of this explicit local alignment, a bi-directional mask recovery module lets each modality reconstruct its masked content under the guidance of the other, implicitly bridging local correspondences into global feature alignment. Furthermore, we construct a new large-scale dataset called T2I-VeRW, which contains 14,668 images covering 1,796 vehicle identities with fine-grained part-level annotations. 
Experimental results on the T2I-VeRI dataset show that PFCVR achieves 29.2\% Rank-1 accuracy, improving over the best competing method by +3.7\% percentage points. On the newly proposed T2I-VeRW benchmark, PFCVR achieves 55.2\% Rank-1 accuracy, outperforming a comprehensive set of recent state-of-the-art methods. 
Source code will be released on \url{https://github.com/Event-AHU/Neuromorphic_ReID} 
\end{abstract}

\begin{IEEEkeywords}
Text-to-Image Vehicle Re-identification, Cross-modal Alignment, Fine-grained Alignment, Vehicle Part Perception 
\end{IEEEkeywords}

\IEEEpeerreviewmaketitle

\section{Introduction}

\IEEEPARstart{W}{ith} the rapid development of the Internet of Things (IoT) \cite{madakam2015internet, javanmardi2021fupe, diro2024differential, khakimov2018iot} in transportation, the Internet of Vehicles (IoV) \cite{dandala2017internet, kadhim2018routing, ang2018deployment} has become an integral part of Intelligent Transportation Systems (ITS)~\cite{song2025text, wang2025knowledge, liu2025knowledge, zhou2022gan, Wu2026VehicleMAEDet, wu2026vehicle, wang2024vehicleMAE, wang2025segmentAV}. The widespread deployment of IoV infrastructure generates large volumes of vehicle image data, substantially advancing vehicle re-identification (Re-ID), the task of matching a query to images from non-overlapping cameras. Current Re-ID methods rely almost exclusively on visual queries, which limits their use when the only available clue is a witness's textual description. In such cases, practitioners must manually locate a matching image from surveillance footage before running the Re-ID pipeline, a process that is labor-intensive and error-prone. Text-based vehicle retrieval accepts a natural-language description as the query, eliminating this manual step and better meeting practical demands.

\begin{figure*}
\centering
\includegraphics[width=\linewidth]{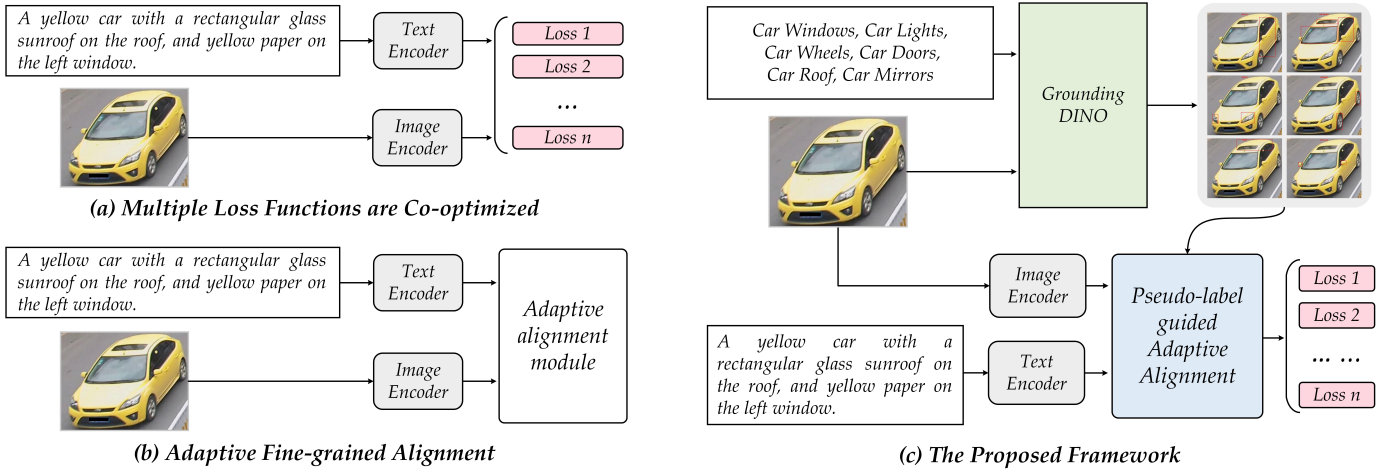}
\caption{Comparison between (a, b) existing text-based vehicle retrieval models and (c) our newly proposed framework.} 
\label{fig:modecompar}
\end{figure*}

The release of VeRI-776~\cite{liu2016deep}, a large-scale multi-view dataset, provided a solid data foundation for vehicle Re-ID. Feng et al.~\cite{feng2021cityflow} extended this line to Text-to-Track Vehicle Retrieval (T2T-VR), which matches text queries to vehicle trajectories in video. However, T2T-VR annotations mainly describe driving state rather than identity-discriminative attributes, and retrieve video clips instead of single images, making it fundamentally different from text-to-image Re-ID. Ding et al.~\cite{ding2024text} then introduced T2I-VeRI, the first text-to-image cross-modal vehicle Re-ID dataset, containing 2,463 image-description pairs across 776 identities built upon VeRI-776. Its descriptions cover fine details, such as sunroof shape, mirror profile, and windshield stickers, making matching substantially more challenging.

Existing methods fall into two paradigms, as shown in Fig.~\ref{fig:modecompar}, the first sub-figure adopts a dual-encoder architecture and co-optimizes multiple losses to align holistic cross-modal features. Representatives such as TIPCB~\cite{chen2022tipcb}, SSAN~\cite{ding2021semantically}, and TFAF~\cite{li2022joint} extract global features through separate encoders supervised by ranking, classification, and matching losses. The second inserts a dedicated alignment module between encoders to capture finer correspondence (as shown in Fig.~\ref{fig:modecompar} (b)): MCANet~\cite{ding2024text} uses three-view masks for multi-level alignment; IRRA~\cite{jiang2023cross} adds an Implicit Relation Reasoning module with cross-attention and masked language modeling; ALBEF~\cite{li2021align} adopts momentum distillation with a matching head.

Despite these advances, two fundamental limitations persist. \textbf{(i) Insufficient part-level alignment.} Existing methods either align features at the holistic level or rely on coarse spatial priors such as the three-view masks in MCANet~\cite{ding2024text}. Such coarse granularity cannot capture fine differences among vehicle components, e.g., headlight shape, mirror style, and roof structure. \textbf{(ii) Disconnect between local and global alignment.} When part-level supervision is introduced, directly aligning isolated part-word embeddings that carry no sentential context conflicts with the global alignment objective, ultimately weakening retrieval accuracy. \textbf{(iii) Existing datasets are limited in scale and insufficient in volume}, making it difficult to meet the data demands of this emerging research direction. The only existing benchmark T2I-VeRI contains merely 2,463 annotated pairs, and this limited scale constrains model training capacity and generalizability.

To overcome these issues, we propose PFCVR (Part-level Fine-grained Cross-modal Vehicle Retrieval), as shown in Fig.~\ref{fig:modecompar} (c). Grounding DINO~\cite{liu2024grounding} detects core vehicle components (windows, wheels, doors, mirrors, lights, and roof) to construct locally paired images and part descriptions as pseudo labels. A Part-level Local Fine-grained Alignment module (PLFA) introduces learnable part-query tokens that aggregate context from both the part text and the full sentence before aligning with visual part features, ensuring that local alignment inherits global context and resolving limitation~(ii). A Bidirectional Mask Recovery Implicit Alignment module (BMRIA) extends unidirectional masked language modeling with Masked Image Modeling, so that each modality recovers its masked content guided by the other, enabling implicit fine-grained alignment without region-level annotation and alleviating limitation~(i). To address data scarcity, we further construct T2I-VeRW, a new large-scale benchmark containing 14,668 images across 1,796 identities with fine-grained part-level annotations.

To sum up, the key contributions of this paper are as follows: 

(1) We propose a novel Part-level Fine-grained Cross-modal Vehicle Retrieval framework (PFCVR), which effectively utilizes locally paired images and texts at the part level. The Part-level Local Fine-grained Alignment Module (PLFA) and Bidirectional Mask Recovery Implicit Alignment Module (BMRIA) are proposed to effectively integrate local alignment and global alignment. 

(2) We construct a new large-scale text-to-image cross-modal vehicle re-identification dataset named T2I-VeRW, which contains 14,668 images with 1,796 vehicle identities and fine-grained part-level annotations. 

(3) Extensive experiments on both T2I-VeRI and T2I-VeRW datasets, including comparisons with recent state-of-the-art methods, fully validate the effectiveness and generalizability of our proposed framework.

\section{Related Work} 
We survey three lines of prior work that bear directly on our approach: Vehicle Re-identification (VReID), Text-to-Track Vehicle Retrieval (T2T-VR), and Text-to-Image Person Re-identification (T2IPReID).

\subsection{Vehicle Re-Identification} 
Vehicle re-identification (VReID) has emerged as a fundamental task in computer vision, and numerous pioneering methods \cite{liu2016large,jeng2013vehicle, liu2017provid, zapletal2016vehicle, liu2016fully, meng2020parsing} have been proposed, achieving notable success and advancing the field. Research on VReID typically focuses on challenges unique to vehicles, which can be broadly categorized into approaches based on vehicle components and those addressing variations in viewing perspectives. Component-based methods~\cite{zheng2021oerff, wang2017orientation} introduce vehicle key points to extract component-level features and jointly optimize global and local representations, but their performance degrades under substantial viewpoint change. An alternative line of work~\cite{Zhou_2018_CVPR, li2021attributes} uses Generative Adversarial Networks (GANs) to synthesize vehicle images from unseen viewpoints, mitigating viewpoint dependency. Nevertheless, when the original viewpoint is already present, GANs struggle to generate realistic images from alternative views, as the synthesized views often deviate significantly from the actual vehicle appearance. To address this issue, several view-aware approaches leverage semantic information to enhance similarity evaluation across different views. For instance, Chen et al. \cite{chen2020orientation} used image-level semantic labels to categorize vehicle images into three viewpoints (front, rear, and side) without requiring manual annotation. A related strategy~\cite{hu2022global} applies view-aware post-processing to refine feature distances within the same view without modifying training. This idea was later extended~\cite{hu2024view} into a more comprehensive framework that systematically models inter-identity feature distances across viewpoint categories, yielding inference-time gains without altering the training pipeline.  

The cross-modal variant of this task was initiated by MCANet~\cite{ding2024text}, which introduced the first text-image vehicle dataset and uses three-view masks for multi-level alignment of shallow, local, and global features. A multimodality adaptive transformer~\cite{zhang2024multimodality} further exploits fine-grained attribute information alongside a mutual learning scheme for unsupervised domain adaptation in vehicle Re-ID, showing that attribute-level cues substantially enhance cross-domain discriminability. However, the reliability 
and granularity of the three-view masks used in MCANet are insufficient for 
fine-grained cross-modal vehicle retrieval. Unlike prior methods, we leverage learnable part queries to explicitly align specific vehicle components with their corresponding textual descriptions, enabling more precise cross-modal matching.

\subsection{Text-to-Track Vehicle Retrieval}
Text-to-Track Vehicle Retrieval (T2T-VR) aims to retrieve a specific vehicle trajectory from a large gallery given a text description. Pioneering methods \cite{khorramshahi2021towards,nguyen2022text,le2022tracked} have achieved strong performance; for instance, Park et al.~\cite{park2021keyword} designed discriminative features to detect relevant vehicles along a trajectory. Because trajectories capture both driving state and interactions with surrounding vehicles, effectively representing them via video clips has become a central challenge. One line of work~\cite{bai2021connecting, zhao2022symmetric} synthesizes a global motion image from the background and vehicle trajectory while using cropped images as local representations, effectively integrating static scene context with dynamic motion patterns. A later extension~\cite{xu2022natural} introduces bounding box sequences to extract local motion features and fuses them with instance, global motion, and segment-level features weighted by importance. Other studies integrate multi-modal cues such as optical flow features~\cite{kim2024moves} to enrich trajectory representations, or augment text descriptions via translation and back-translation~\cite{bai2021connecting} and NLP-based query expansion~\cite{zhang2022multi} to improve robustness~\cite{fadaeddini2023data}.  

Recent work has further optimized matching strategies through contrastive learning and attention mechanisms. The success of CLIP~\cite{radford2021learning} in cross-modal tasks has also inspired adaptation to T2T-VR; for example, Zhu et al.~\cite{zhu2024vision} combined CLIP with interactive prompt learning for multimodal tracking in ITS scenarios, achieving robust cross-modal alignment with minimal additional parameters. Unlike trajectory-level retrieval, our task pairs single static images with witness descriptions, which calls for part-level visual grounding rather than motion or context modeling. We draw on the T2T-VR literature primarily for insight into contrastive training strategies and the use of pre-trained multimodal backbones.

\begin{figure*}
\centering
\includegraphics[width=0.95\linewidth]{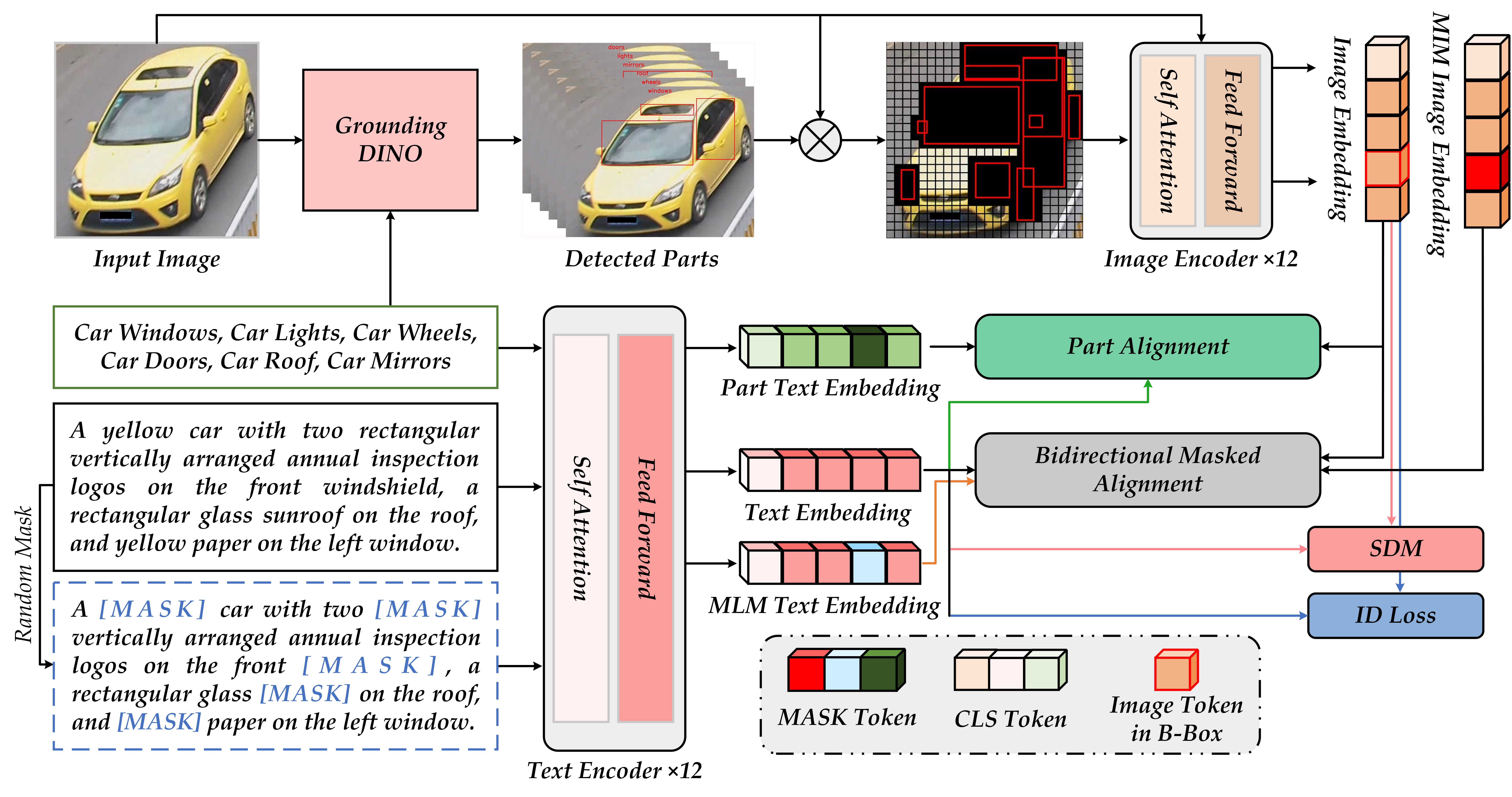}
\caption{An illustration of our proposed part-level alignment framework based on Grounding DINO for cross-modal vehicle retrieval.} 
\label{fig:Framework}
\end{figure*}

\subsection{Text-to-Image Person Re-Identification}

As another fine-grained task in cross-modal retrieval, Text-to-Image Person Re-identification (T2IPReID) aims to rank pedestrian images in a gallery according to a given textual description and retrieve the most relevant samples \cite{li2017person}. Existing T2IPReID methods can be broadly divided into two categories based on their alignment granularity: global matching methods \cite{zhu2021dssl,shu2022see} and local matching methods \cite{shao2022learning,wang2021text}. Global matching approaches typically employ losses such as CMPM/C loss \cite{zhang2018deep}, Triplet Ranking loss \cite{faghri2017vse++}, and SDM loss \cite{jiang2023cross} to align the global features of both modalities in a shared latent space. While effective, these methods focus primarily on holistic features and neglect fine-grained alignment between local regions, thereby limiting performance.  

To address this limitation, local matching methods explicitly explore region-level associations between body parts and textual entities. TIPCB~\cite{chen2022tipcb}, for example, adopts a multi-stage matching strategy that incorporates local, global, and shallow features to reduce inter-modal discrepancies. Pose-guided methods~\cite{jing2020pose} leverage human pose keypoints to guide visual local feature extraction. However, such approaches generally demand greater computational resources due to the complexity of modeling local correspondences.  

Moreover, given the weak correspondence between text and images, some studies investigate alignment between unpaired samples. Ding et al. \cite{ding2021semantically} introduced Composite Ranking (CR) 
loss to constrain the alignment of unpaired samples of images and texts from the same 
pedestrian, thereby mining additional positive samples. More recently, 
Xiao et al. \cite{xiao2025reid} proposed to fuse semantic and attribute 
information into a unified person re-identification network for ITS 
applications, showing that jointly leveraging fine-grained semantic cues 
with identity-level supervision yields stronger and more generalizable 
cross-modal representations.

In parallel, several 2025 methods have pushed T2IPReID accuracy to new levels. VFE-TPS~\cite{shen2025enhancing} pairs a Text Guided Masked Image Modeling task with an Identity Supervised Global Visual Feature Calibration task to transfer CLIP knowledge for local visual detail capture without explicit part annotation. Rather than full fine-tuning, UP-Person~\cite{liu2025up} unifies three lightweight parameter-efficient modules (Prefix, LoRA, and Adapter), achieving competitive accuracy while updating only 4.7\% of backbone parameters. FMFA~\cite{yin2025cross} targets the misaligned-positive-pair problem through Adaptive Similarity Distribution Matching, complemented by an Explicit Fine-grained Alignment module that sparsifies the similarity matrix for hard-coded local matching. To compensate for limited person-centric data, GA-DMS~\cite{zheng2025gradient} curates WebPerson, a 5M-pair web-sourced dataset, and applies Gradient-Attention guided Dual-Masking to suppress noisy text tokens during contrastive learning. MARS~\cite{ergasti2025mars} combines a Visual Reconstruction Loss, in which a Masked AutoEncoder reconstructs image patches conditioned on text, with an Attribute Loss that reweights adjective-noun attribute chunks.

Despite these advances, all five methods are designed for pedestrian retrieval and share two shortcomings when transferred to the vehicle domain. First, they rely on human-body structural priors (pose keypoints, body-part parsing, or pedestrian attribute taxonomies) that do not translate to the rigid, symmetric, and view-dependent geometry of vehicles. Second, their alignment mechanisms operate at either the holistic level or the implicit feature level, without explicit part-level supervision tailored to vehicle-specific components such as headlights, mirrors, and roof structures. Our PFCVR addresses both gaps: PLFA provides explicit part-level alignment via learnable part-query tokens grounded in vehicle component detection, and BMRIA couples masked language and image modeling for bidirectional implicit reasoning adapted to vehicle part geometry.

\section{Our Proposed Approach}

The overall architecture of PFCVR is introduced in
Section~\ref{Overview}.  The Part-level Local Fine-grained Alignment
Module and the Bidirectional Mask Recovery Implicit Alignment Module are
described in Sections~\ref{PLFA} and~\ref{BMRIA}, respectively.
Section~\ref{DataAug} covers the data augmentation strategy and
Section~\ref{loss} defines the training objectives.

\subsection{Overview\label{Overview}}

As illustrated in Fig.~\ref{fig:Framework}, PFCVR takes a vehicle
image and its paired textual description as input. Grounding
DINO~\cite{liu2024grounding} first localizes six predefined vehicle
components (windows, wheels, doors, mirrors, lights, and roof)
by grounding their corresponding part-level keywords in the image.
Two augmented inputs are derived from this detection step: a
\emph{part-masked image}, constructed by zeroing out all detected part
regions in the original image, and a \emph{masked text}, obtained by
randomly masking tokens in the textual description.

Five inputs are then encoded by CLIP's pre-trained vision encoder (ViT-B/16)
and text encoder: the original image, the original text, the part-masked
image, the masked text, and per-part textual descriptions.  These
produce the global image feature $F_G^{\mathcal{I}}$, global text
feature $F_G^{T}$, masked image feature $F_M^{\mathcal{I}}$, masked text
feature $F_M^{T}$, and part-level text feature set
$\{F_{P_k}^{T}\}_{k=1}^{K}$, where $K=6$ is the number of predefined part
categories (windows, wheels, doors, mirrors, lights, and roof).

Two lightweight modules are stacked on top of the encoder.  The PLFA
module (Section~\ref{PLFA}) introduces learnable part-query tokens that
first absorb context from the full sentence before aligning with visual
part features, eliminating the representation mismatch that arises from
using isolated part-word embeddings.  The BMRIA module
(Section~\ref{BMRIA}) drives implicit fine-grained alignment through
bidirectional masked reconstruction: masked image patches are recovered
with the aid of text features, and masked text tokens are recovered with
the aid of image features.  The full training objective combines SDM
loss and ID loss for global alignment with the part-level ITC loss from
PLFA and the bidirectional reconstruction loss from BMRIA.

\begin{figure*}
\centering
\includegraphics[width=\linewidth]{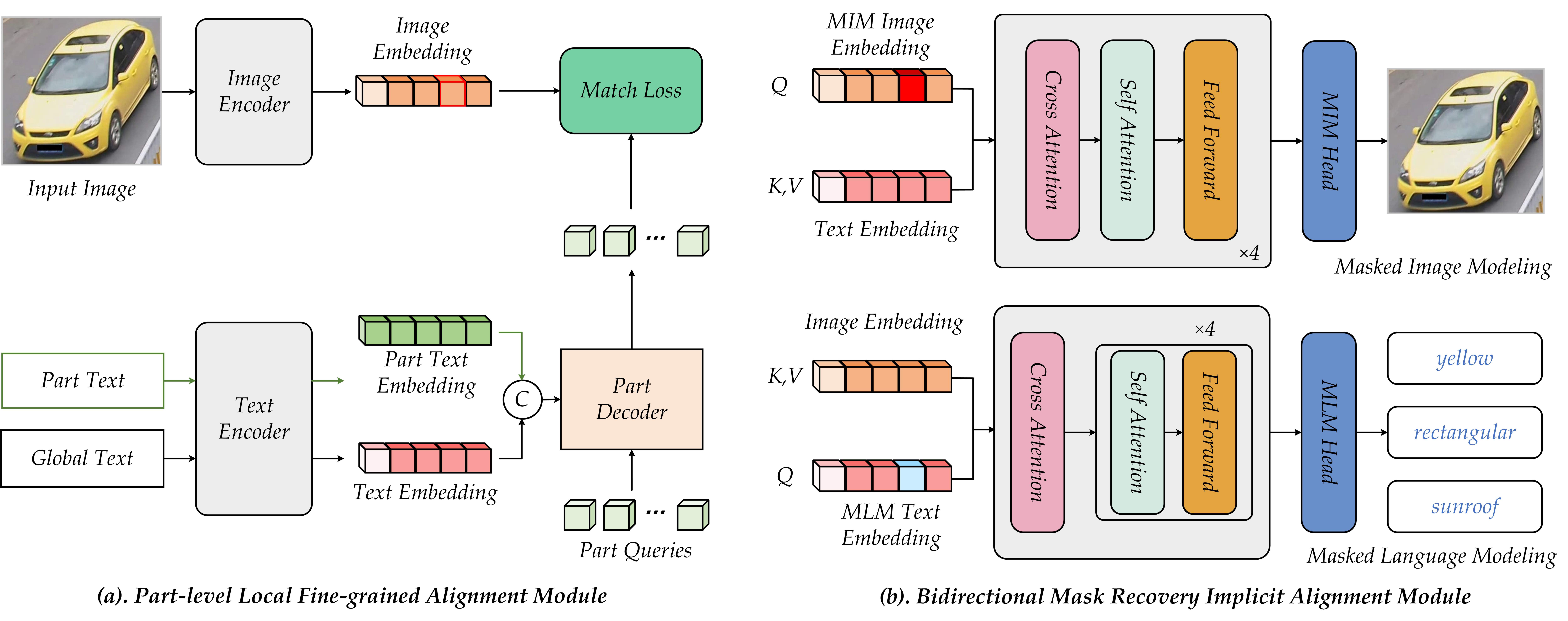}
\caption{Detailed network architecture of the proposed (a) Part-level Local Fine-grained Alignment (PLFA) Module and (b) Bidirectional Mask Recovery Implicit Alignment (BMRIA) Module.}  
\label{fig:LocalModules}
\end{figure*}

\subsection{Part-level Local Fine-grained Alignment Module\label{PLFA}}

A direct way to exploit part-level supervision is to align the
bounding-box image features obtained from Grounding DINO with the
embedding of the corresponding part keyword (\eg, aligning the region
inside a ``car door'' box with the word vector of ``door'').  However,
isolated part-word embeddings carry no context from the surrounding
description, creating a representation gap that causes the local
alignment objective to conflict with the global alignment across the two modalities
and ultimately degrade retrieval performance.

To close this gap, we introduce a set of learnable part-query tokens
$Q_P \in \mathbb{R}^{K \times d}$, one per part category, inspired by
the object query design in DETR~\cite{carion2020end}.  As illustrated in
Fig.~\ref{fig:LocalModules}(a), each token aggregates context from both
the standalone part text $F_P^{T}$ and the full-sentence global text
$F_G^{T}$ via multi-head cross-attention:

\begin{equation}
  Q_P \leftarrow \mathrm{MCA}\!\left(
      \mathrm{Proj}(Q_P),\;
      \mathrm{Concat}\!\left(F_P^{T},\, F_G^{T}\right)
  \right),
  \label{eq:plfa_query}
\end{equation}
where $\mathrm{Proj}(\cdot)$ is a linear projection applied to the
queries, $\mathrm{Concat}(\cdot)$ concatenates the part and global text
features along the sequence dimension, and MCA denotes multi-head
cross-attention.  After this update, $Q_P$ encodes part-specific
semantics that remain coherent with the global sentence, resolving the
representation mismatch noted above.

The corresponding visual part features are extracted by masking the
global image feature map with the part detection mask
$M_P \in \{0,1\}^{K \times L_v}$, where $L_v$ is the number of visual
patch tokens.  For each part $k$, the mask entry is set to 1 for patches
that fall within the corresponding Grounding DINO bounding box and 0
otherwise:

\begin{equation}
  \begin{aligned}
    F_P^{\mathcal{I}} &= M_P \odot F_G^{\mathcal{I}}, \\
    \mathcal{L}_{\mathrm{ITC}} &= \mathrm{ITC}\!\left(
        F_P^{\mathcal{I}},\; Q_P
    \right),
  \end{aligned}
  \label{eq:plfa_align}
\end{equation}
where $\odot$ denotes element-wise masking and $\mathrm{ITC}(\cdot)$ is
the contrastive loss defined in Section~\ref{loss}.
By aligning $F_P^{\mathcal{I}}$ against $Q_P$ rather than raw part-word
embeddings, the local supervision signal remains consistent with
the global alignment objective.

\subsection{Bidirectional Mask Recovery Implicit Alignment
Module\label{BMRIA}}

PLFA provides explicit supervision at the part level, but each part is
treated as an independent alignment target.  BMRIA addresses this gap
from the global side: each modality is partially masked and then
reconstructed using features from the other.
Because the cross-attention in each reconstruction branch must locate
the relevant counterpart regions to perform accurate recovery, the model
is implicitly trained to build fine-grained cross-modal correspondences
without any explicit region-token annotation.  Fig.~\ref{fig:LocalModules}(b)
illustrates both branches.

\noindent\textbf{Image-guided text reconstruction.}
The masked text feature $F_M^{T}$ is decoded by attending to the global
image feature $F_G^{\mathcal{I}}$:

\begin{equation}
  \begin{aligned}
    \tilde{F}_R^{T} &= \mathrm{MSA}\!\left(
        \mathrm{FFN}\!\left(
            \mathrm{MCA}\!\left(F_M^{T},\, F_G^{\mathcal{I}}\right)
        \right)
    \right), \\
    \hat{T} &= \mathrm{Head}_{\mathrm{MLM}}\!\left(\tilde{F}_R^{T}\right),
  \end{aligned}
  \label{eq:bmria_text}
\end{equation}
where MSA is multi-head self-attention, FFN is a feed-forward network,
and $\mathrm{Head}_{\mathrm{MLM}}$ is a linear classifier over the
vocabulary.  Text tokens are semantically compact and discrete, so a
single MCA layer is sufficient to integrate the image context.

\noindent\textbf{Text-guided image reconstruction.}
The masked image feature $F_M^{\mathcal{I}}$ is decoded by attending to
the global text feature $F_G^{T}$:

\begin{equation}
  \begin{aligned}
    \tilde{F}_R^{\mathcal{I}} &= \mathrm{MSA}\!\left(
        \mathrm{FFN}\!\left(
            \mathrm{MCA}\!\left(F_M^{\mathcal{I}},\, F_G^{T}\right)
        \right)
    \right), \\
    \hat{\mathcal{I}} &= \mathrm{Head}_{\mathrm{MIM}}\!\left(
        \tilde{F}_R^{\mathcal{I}}
    \right),
  \end{aligned}
  \label{eq:bmria_image}
\end{equation}
where $\mathrm{Head}_{\mathrm{MIM}}$ projects each masked patch token
back to pixel space.  Pixel-level recovery from inherently sparse text
features is a harder task than token classification; we therefore stack
multiple MCA layers in this branch to provide sufficient cross-modal
capacity.  The reconstruction losses for both branches are defined in
Section~\ref{loss}.

\subsection{Data Augmentation\label{DataAug}}

As shown in Table~\ref{tab:distribution}, most vehicle identities in
T2I-VeRI are associated with only two or three annotated pairs,
leaving insufficient data for the model to learn discriminative
representations.  To address this imbalance, we apply two augmentation
operations to each training image, generating three additional copies
per original sample and expanding the training set to four times its
original size.

\emph{1) Brightness adjustment:} Two variants are produced by modifying
the Gamma value of the image, yielding one over-exposed and one
under-exposed copy that simulate the photometric variation caused by
differences in camera installation and lighting conditions.

\emph{2) Gaussian noise injection:} One variant is produced by adding
randomly sampled Gaussian noise to the original image, simulating
sensor-level interference present in real surveillance footage.

Each augmented image is paired with the original textual description to
form a new training pair.  No augmentation is applied to the
test set.





\subsection{Loss Function\label{loss}}

The overall loss function used in this work can be summarized as: 
\begin{equation}
  \mathcal{L} =
    \alpha\,\mathcal{L}_{\mathrm{ID}}
  + \beta\,\mathcal{L}_{\mathrm{SDM}}
  + \gamma\,\mathcal{L}_{\mathrm{ITC}}
  + \delta\,\mathcal{L}_{\mathrm{BiIRR}},
\end{equation}
where $\alpha, \beta, \gamma, \delta$ are set to $0.5, 1.0, 0.2, 0.5$, respectively. 
The detailed introduction to these items is given below.

\noindent\textbf{ID Loss.}
The identity classification loss encourages the model to produce
class-discriminative embeddings.  A shared linear classifier
$\mathrm{Head}_{\mathrm{ID}}$ projects global image and text features
onto identity logits:

\begin{equation}
  \begin{aligned}
    P_{\mathcal{I}} &= \mathrm{Softmax}\!\left(
        \mathrm{Head}_{\mathrm{ID}}(F_G^{\mathcal{I}})
    \right), \\
    P_{T} &= \mathrm{Softmax}\!\left(
        \mathrm{Head}_{\mathrm{ID}}(F_G^{T})
    \right).
  \end{aligned}
\end{equation}
The loss is the average cross-entropy over both modalities:
\begin{equation}
  \mathcal{L}_{\mathrm{ID}} =
    -\frac{1}{2}\!\sum_{m \in \{\mathcal{I},\, T\}}
     \sum_{i=1}^{N} \mathbf{y}_i \log P_m^{(i)},
\end{equation}
Here, $N$ is the batch size and $\mathbf{y}_i$ the one-hot identity label of sample $i$.

\noindent\textbf{SDM Loss.}
The Similarity Distribution Matching loss~\cite{jiang2023cross} aligns
the pairwise similarity distribution across modalities.  For a batch of
$N$ paired samples with cosine similarity matrix
$S \in \mathbb{R}^{N \times N}$:
\begin{equation}
  \begin{aligned}
    P_{ij} &= \frac{\exp(S_{ij}/\tau)}{\sum_{k}\exp(S_{ik}/\tau)}, \\
    \mathcal{L}_{\mathrm{SDM}} &= \frac{1}{2}\!\left(
        \mathrm{KL}(P_{\mathcal{I}\to T} \| Q)
      + \mathrm{KL}(P_{T \to \mathcal{I}} \| Q)
    \right),
  \end{aligned}
\end{equation}
The temperature scalar $\tau$ controls the sharpness of the distribution.  $P_{\mathcal{I}\to T}$ is the row-wise softmax over the similarity matrix (image-to-text direction), $P_{T \to \mathcal{I}}$ its column-wise counterpart, and $Q_{ij} = \mathbb{1}[y_i = y_j] / \sum_k \mathbb{1}[y_i = y_k]$ the ground-truth identity distribution.  Because SDM matches soft distributions rather than requiring hard positives, it stabilizes training on the small T2I-VeRI dataset.

\noindent\textbf{ITC Loss.}
The contrastive loss between images and texts~\cite{radford2021learning} is used in
PLFA to align part-level image features $F_P^{\mathcal{I}}$ with
part-query tokens $Q_P$:
\begin{equation} 
\small 
\mathcal{L}_{\mathrm{ITC}} = -\frac{1}{2N}
    \sum_{i=1}^{N}\!\left[
      \log\frac{\exp(S_{ii}^{\mathrm{P}}/\tau)}{\sum_j \exp(S_{ij}^{\mathrm{P}}/\tau)}
    + \log\frac{\exp(S_{ii}^{\mathrm{P}}/\tau)}{\sum_j \exp(S_{ji}^{\mathrm{P}}/\tau)}
    \right],
\end{equation}
where $S_{ij}^{\mathrm{P}}$ is the cosine similarity between the $i$-th part image
feature and the $j$-th part-query feature.

\noindent\textbf{BiIRR Loss.}
The Bidirectional Implicit Relation Reasoning loss supervises the two
reconstruction branches in BMRIA.  The \emph{image-guided text
reconstruction} branch is trained with masked-language-modeling
cross-entropy:
\begin{equation}
  \mathcal{L}_{\mathrm{BiIRR}}^{i \to t} =
    -\frac{1}{|M_T|}
     \sum_{p \in M_T}
     \log \frac{\exp(m_p^{y_p})}{\sum_{k=1}^{|V|} \exp(m_p^{k})},
\end{equation}
Above, $M_T$ collects the masked token positions, $V$ is the vocabulary,
$m_p^{k}$ the predicted logit for vocabulary entry $k$ at position $p$,
and $y_p$ the ground-truth token index.
The \emph{text-guided image reconstruction} branch follows the
MAE~\cite{he2022masked} objective with pixel-level MSE:
\begin{equation}
  \mathcal{L}_{\mathrm{BiIRR}}^{t \to i} =
    \frac{1}{|M_{\mathcal{I}}|}
    \sum_{j \in M_{\mathcal{I}}} \!\left(x_j - \hat{x}_j\right)^2,
\end{equation}
where $M_{\mathcal{I}}$ is the set of masked image patch positions,
$x_j$ is the ground-truth patch, and $\hat{x}_j$ is the reconstructed
patch.  The combined BiIRR loss is:
\begin{equation}
  \mathcal{L}_{\mathrm{BiIRR}} =
    \frac{\mathcal{L}_{\mathrm{BiIRR}}^{i \to t}
        + \mathcal{L}_{\mathrm{BiIRR}}^{t \to i}}{2}.
\end{equation}

\begin{table}
\center
\small 
\caption{Distribution of data in T2I Vehicle ReID dataset. Each column header indicates the number of annotated pairs per identity, and the cell value indicates how many IDs fall into that group.} 
\label{tab:distribution}
\begin{tabular}{c|cccc}
\hline \toprule [0.5 pt] 
  \multicolumn{1}{c|}{Dataset partition} &
  \multicolumn{1}{c}{2} & 
  \multicolumn{1}{c}{3} &
  \multicolumn{1}{c}{4}&
  \multicolumn{1}{c}{5} \\ \hline
 All & 149 & 361 & 248  & 18  \\
 Train & 106 & 251 & 176  & 13  \\
 Test & 43 & 110 & 72  & 5  \\
\hline \toprule [0.5 pt] 
\end{tabular} 
\end{table}

\begin{figure}
\centering
\includegraphics[width=\linewidth]{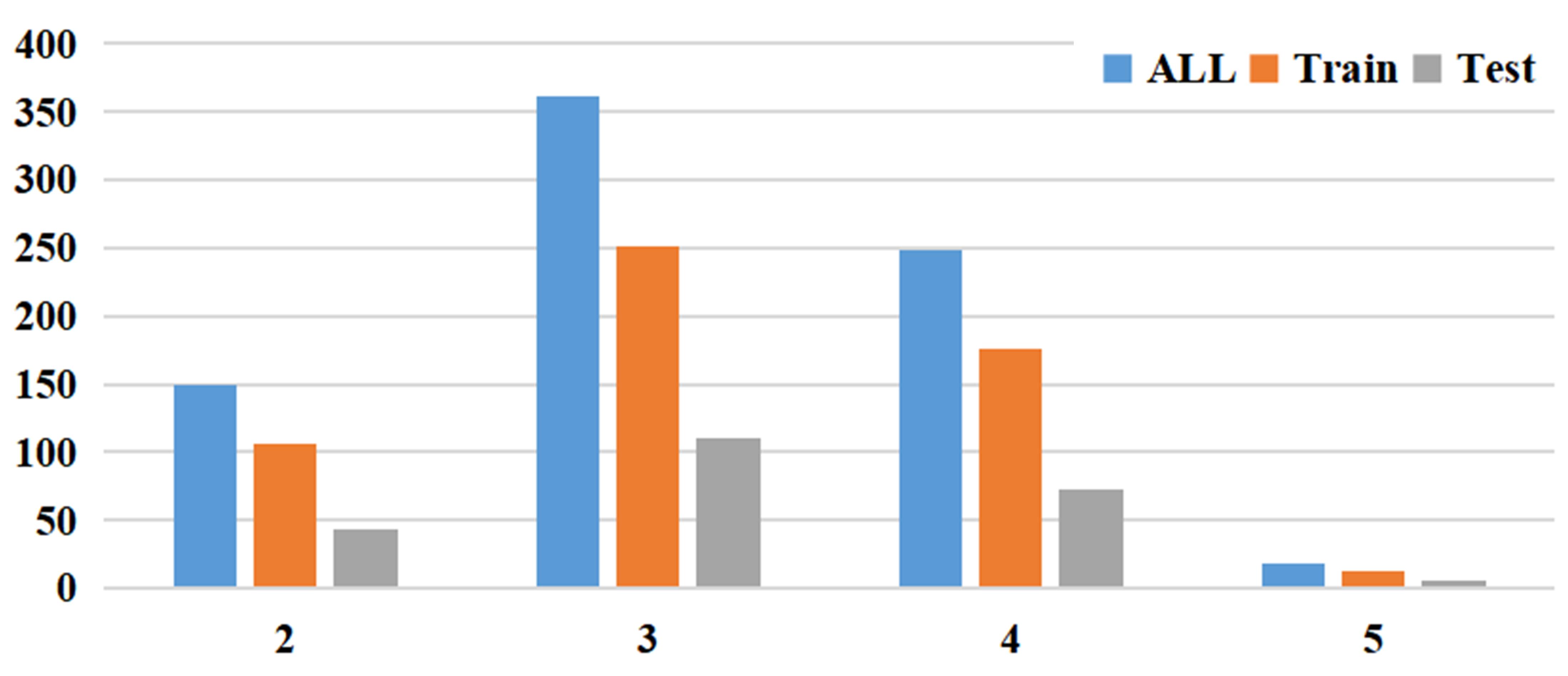}
\caption{Distribution of data in T2I Vehicle ReID dataset. The horizontal axis indicates the number of annotated pairs per identity, and the vertical axis indicates how many identities fall into each group.} 
\label{fig:distribution}
\end{figure}

\begin{figure*}
\centering
\includegraphics[width=\linewidth]{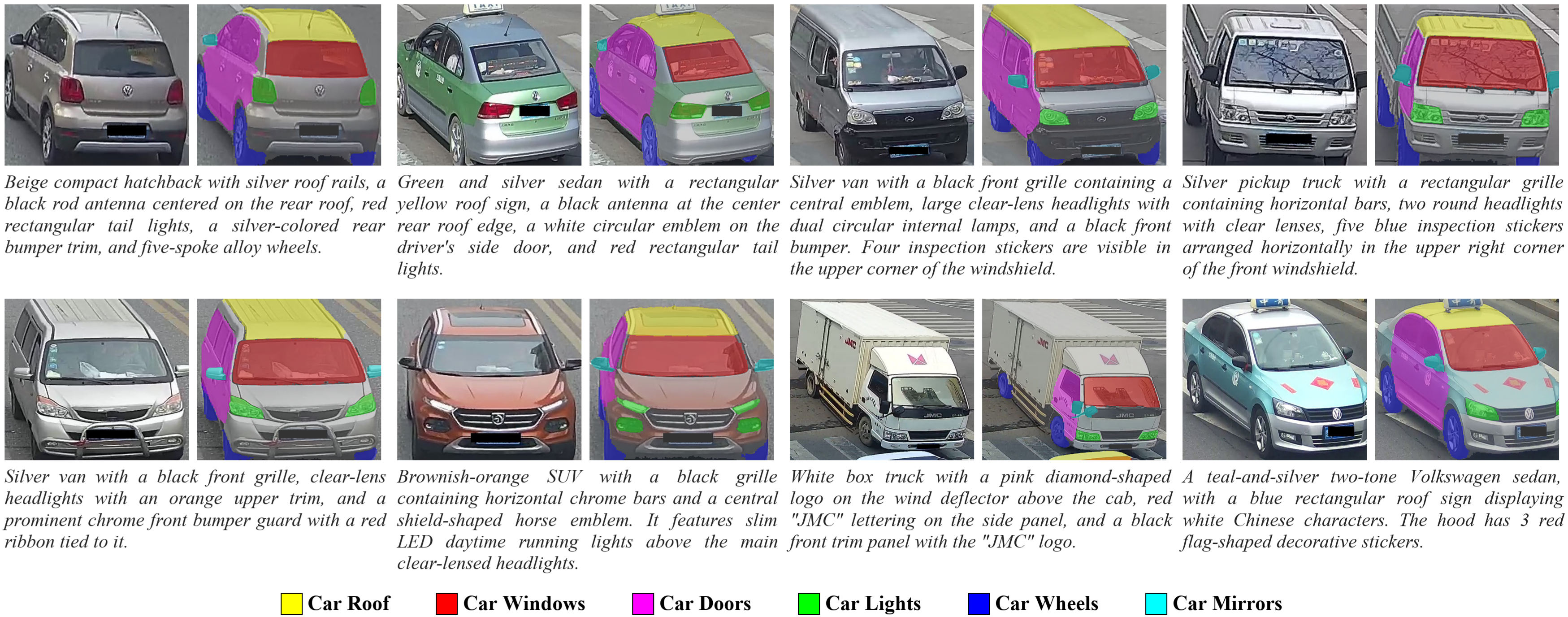}
\caption{Representative samples from the T2I-VeRW dataset. Each sample shows the original vehicle image (left) and its part-level segmentation overlay (right), along with the corresponding textual description.}
\label{fig:dataset_visualization}
\end{figure*}

\section{T2I-VeRW Benchmark Dataset}\label{sec:t2iverw} 

To address the data scarcity that limits existing cross-modal vehicle re-identification research, we construct T2I-VeRW, a new large-scale benchmark derived from the VERI-Wild dataset. It contains 14,668 vehicle images covering 1,796 distinct vehicle identities, substantially exceeding the scale and diversity of T2I-VeRI.

\subsection{Protocols and Data Annotation} 

The dataset is divided into training and validation sets with a 7:3 split ratio, ensuring that no vehicle identities overlap between the two subsets. The training set contains 10,265 annotated pairs from 1,251 vehicle identities, while the validation set contains 4,403 annotated pairs from the remaining 545 identities.

For text annotation, we employ the Qwen3-VL-32B~\cite{bai2025qwen3} multimodal large language model with a carefully designed prompt template that explicitly specifies key vehicle attributes, including grille style, headlight shape, wheel design, roof features, window details, and bumper characteristics. The generated descriptions are subsequently reviewed and manually refined to ensure linguistic fluency and semantic accuracy. As a result, the captions have an average length of 50.35 words, which is significantly longer than the average caption length in T2I-VeRI (27.6 words). This richer linguistic detail simultaneously poses a more demanding alignment challenge and provides stronger part-level supervision signals.

For part-level annotations, we employ SAM3 (Segment Anything Model 3)~\cite{carion2025sam3} to generate segmentation masks for six vehicle parts: windows, lights, wheels, roof, doors, and mirrors. Each part category is segmented using multiple textual prompts (e.g., ``car windows, windshield, side windows, rear window'' for the window category) to improve coverage and robustness. The resulting masks are stored as $24\times24$ binary matrices and verified through random visualization sampling.

\subsection{Statistical Analysis}

Table~\ref{tab:dataset_statistics} compares T2I-VeRI and T2I-VeRW. T2I-VeRW is nearly six times larger than T2I-VeRI in image count and contains more than twice as many identities. The longer average caption (50.35 vs.\ 27.6 words) reflects finer annotation granularity, covering attributes such as body color, roof sign, tail light shape, wheel spoke pattern, and sticker placement.

\begin{table}[!htb]
\center
\small 
\caption{Statistics of T2I-VeRI and T2I-VeRW datasets.} 
\label{tab:dataset_statistics}
\begin{tabular}{c|cc}
\hline \toprule [0.5 pt] 
  \multicolumn{1}{c|}{\textbf{Statistics}} &
  \multicolumn{1}{c}{\textbf{T2I-VeRI}} & 
  \multicolumn{1}{c}{\textbf{T2I-VeRW}} \\ \hline
 Total Images & 2,463 & 14,668 \\
 Total IDs & 776 & 1,796 \\
 Training Images & 1,734 & 10,265 \\
 Training IDs & 546 & 1,251 \\
 Test/Val Images & 729 & 4,403 \\
 Test/Val IDs & 230 & 545 \\
 Part Categories & 6 & 6 \\
 Avg. Words per Caption & 27.6 & 50.35 \\
\hline \toprule [0.5 pt] 
\end{tabular} 
\end{table}

Figure~\ref{fig:dataset_visualization} presents representative samples from T2I-VeRW, illustrating the diversity of vehicle types, viewpoints, and the quality of both textual descriptions and part-level segmentation masks. Each sample shows the original image alongside its part segmentation overlay, where six categories (roof, windows, doors, lights, wheels, and mirrors) are color-coded.

\subsection{Benchmark Construction}

The part annotation strategy used for T2I-VeRW differs from the pipeline adopted for T2I-VeRI in our framework. For T2I-VeRI, we employ Grounding DINO to obtain bounding boxes of vehicle parts. In contrast, for T2I-VeRW, we leverage SAM3 to generate pixel-level segmentation masks, which are subsequently discretized into $24\times24$ grids. This SAM-based strategy provides more precise part-level representations than bounding-box annotations.

The evaluation protocol follows the T2I-VeRI benchmark; detailed metric definitions are given in Section~\ref{datasetandmetric}. The non-overlapping identity split and the richer annotation granularity make T2I-VeRW a more challenging and comprehensive testbed for cross-modal vehicle retrieval methods.

\section{Experiments} 
We evaluate PFCVR on both T2I-VeRI and T2I-VeRW. The datasets, evaluation metrics, and implementation details are described first, followed by comparisons with state-of-the-art methods, ablation studies, and a visual analysis of retrieval results.

\subsection{Datasets and Evaluation Metrics\label{datasetandmetric}} 

\textbf{T2I-VeRI dataset.} The T2I-VeRI dataset \cite{ding2024text} is the first publicly available benchmark for text-to-image cross-modal vehicle re-identification. It contains 2,463 annotated pairs of images and descriptions corresponding to 776 unique vehicle identities. Following the official split protocol \cite{ding2024text}, the dataset is divided into non-overlapping training and test sets. The training set consists of 1,734 paired samples from 546 vehicle identities, while the test set includes 729 paired samples from the remaining 230 identities.

\textbf{T2I-VeRW dataset.} See Section~\ref{sec:t2iverw} for construction details and Table~\ref{tab:dataset_statistics} for a side-by-side comparison with T2I-VeRI.

\textbf{Evaluation Metrics.} Rank-$k$ accuracy ($k=1,5,10$) and mean Average Precision (mAP), which are widely used in Re-ID and cross-modal retrieval tasks, serve as the evaluation criteria. Rank-$k$ accuracy measures the probability that a correct match appears within the top $k$ retrieved results. The formal definition is:




\begin{equation}
\begin{split}
\text{Rank-}k = \frac{Q_{\text{success}}}{Q_{\text{all}}}
\end{split}
\end{equation}
with $Q_{\text{success}}$ being the number of correct matches within the top $k$ results and $Q_{\text{all}}$ the total number of queries.  

Mean Average Precision (mAP) jointly considers precision and recall by computing the average precision for each query and then averaging across all queries:



\begin{equation}  
\begin{aligned} 
&mAP = \frac{1}{Q_{all}}\sum_{q=1}^{Q_{all}}AP_{q},\\
&AP_q = \frac{1}{m_q}\sum_{k=1}^{Q_{all}}P_q(k)\cdot \theta_q(k),\\
&P_q(k) =\frac{\Theta_q (k)}{k} 
\end{aligned}
\end{equation}
where $Q_{all}$ represents the total number of queries. For a given query $q$, $m_q$ is the total number of ground-truth matches in the test set. $\Theta_q(k)$ denotes the count of correct matches within the top $k$ retrieved results. $\theta_q(k)$ is an indicator function, which is 1 if the $k$-th retrieved result is a correct match and 0 otherwise. 

\subsection{Implementation Details\label{implementation}}

Our proposed PFCVR consists of a pre-trained image encoder (CLIP-ViT-B/16), a pre-trained text encoder (CLIP Text Transformer), and two randomly initialized interactive encoders for text-to-image and image-to-text feature fusion, respectively. In each layer of the interactive encoders, the hidden dimension is set to 512, with 8 attention heads.
During training, several data augmentation strategies are applied, including random horizontal flipping, random cropping, and random erasing. All input images are resized to $384 \times 384$. The maximum token length of the text sequence, denoted as $L$, is set to 77. In the Bidirectional Mask Recovery Implicit Alignment (BMRIA) module, the mask ratios for the visual and textual branches are set to 0.25 and 0.15, respectively.

The model is trained for 60 epochs using the Adam optimizer \cite{kingma2014adam}. The initial learning rate is set to $1\times10^{-5}$ and follows a cosine decay schedule. During the first five warm-up epochs, the learning rate is linearly increased from $1\times10^{-6}$ to $1\times10^{-5}$. For the randomly initialized modules, a larger initial learning rate of $5\times10^{-5}$ is adopted. The loss weights are set as follows: $\alpha=0.5$ for ID Loss, $\beta=1.0$ for SDM Loss, $\gamma=0.2$ for ITC Loss, and $\delta=0.5$ for BiIRR Loss. All experiments are conducted on a single NVIDIA RTX 4090 GPU with 24 GB memory. More details can be found in our source code.

\begin{table}
\center
\small 
\caption{Comparison with state-of-the-art methods on the T2I-VeRI dataset. Best results in red, second best in blue.
} 
\label{quantitativeresults}
\resizebox{\columnwidth}{!}{
\begin{tabular}{c|cc|cccc}
\hline \toprule [0.5 pt] 
\multicolumn{1}{c|}{\multirow{2}{*}{\textbf{\makecell[c]{Methods}}}} & \multicolumn{1}{c}{\multirow{2}{*}{\textbf{Venue}}} & \multicolumn{1}{c|}{\multirow{2}{*}{\textbf{Year}}} & \multicolumn{4}{c}{\textbf{T2I-VeRI \cite{ding2024text}}}  \\ \cline{4-7} 

  \multicolumn{1}{c|}{} &
  \multicolumn{1}{c}{} &
  \multicolumn{1}{c|}{} &
  \multicolumn{1}{c}{Rank-1} & 
  \multicolumn{1}{c}{Rank-5} &
  \multicolumn{1}{c}{Rank-10}&
  \multicolumn{1}{c}{mAP} \\ \hline
 TIPCB \cite{chen2022tipcb} & Neurocomputing & 2022 & 16.8 & 44.0 &58.8  & 18.0  \\
 LGUR \cite{shao2022learning} & ACM MM & 2022 & 11.3 & 29.0 &45.9  & 9.2  \\
 SSAN \cite{ding2021semantically} & arXiv & 2021 & 14.2 & 34.3 &52.6  & 12.2  \\
 TFAF \cite{li2022joint} & IEEE SPL & 2022 & 20.1 & 49.0 &66.3  & 15.5  \\
 TransReID \cite{he2021transreid} & ICCV & 2021 & 7.5 & 24.3 &35.1  & 7.4  \\
 HAT \cite{bin2023unifying} & ACM MM & 2023 & 16.0 & 40.7 &56.2  & 13.7  \\
 ALBEF \cite{li2021align} & NeurIPS & 2021 & 22.9 & 49.6 &66.1  & 21.8  \\
 MCANet \cite{ding2024text} & IEEE T-ITS & 2024 & 25.1 & 54.7 &69.1  & 18.1  \\
 IRRA \cite{jiang2023cross} & CVPR & 2023 & 25.1 &  \textcolor{blue}{57.8} & \textcolor{blue}{72.4}  & 23.7  \\
 UP-Person \cite{liu2025up} & TCSVT & 2025 & 18.8 & 46.4 & 61.3 & 18.5 \\
 FMFA \cite{yin2025cross} & ACM TOMM & 2025 & 21.8 & 53.6 & 69.1 & 21.6 \\
 GA-DMS \cite{zheng2025gradient} & EMNLP & 2025 & 22.2 & 53.6 & 69.0 & 21.5 \\
 VFE-TPS \cite{shen2025enhancing} & KBS & 2025 & 22.2 & 53.1 & 69.4 & 21.7 \\
 MARS \cite{ergasti2025mars} & ACM TOMM & 2025 & \textcolor{blue}{25.5} & 54.3 & 67.6 & \textcolor{blue}{24.9} \\
PFCVR (Ours) & - & - & \textcolor{red}{29.2} & \textcolor{red}{60.1} &\textcolor{red}{75.4}  & \textcolor{red}{25.3}  \\
\hline \toprule [0.5 pt] 
\end{tabular} 
}
\end{table}

\begin{table}
\center
\small 
\caption{Comparison with state-of-the-art methods on the T2I-VeRW dataset. Best results in red, second best in blue.
} 
\label{quantitativeresults_new}
\resizebox{\columnwidth}{!}{
\begin{tabular}{c|cc|cccc}
\hline \toprule [0.5 pt] 
\multicolumn{1}{c|}{\multirow{2}{*}{\textbf{\makecell[c]{Methods}}}} & \multicolumn{1}{c}{\multirow{2}{*}{\textbf{Venue}}} & \multicolumn{1}{c|}{\multirow{2}{*}{\textbf{Year}}} & \multicolumn{4}{c}{\textbf{T2I-VeRW}}  \\ \cline{4-7} 
  \multicolumn{1}{c|}{} &
  \multicolumn{1}{c}{} &
  \multicolumn{1}{c|}{} &
  \multicolumn{1}{c}{Rank-1} & 
  \multicolumn{1}{c}{Rank-5} &
  \multicolumn{1}{c}{Rank-10} &
  \multicolumn{1}{c}{mAP} \\ \hline
 TIPCB \cite{chen2022tipcb} & Neurocomputing & 2022 & 29.1 & 53.3 & 64.5 & 17.2  \\
 LGUR \cite{shao2022learning} & ACM MM & 2022 & 44.2 & 73.0 & 83.1 & 18.5  \\
 SSAN \cite{ding2021semantically} & arXiv & 2021 & 39.1 & 67.3 & 78.6 & 21.6  \\
 TransReID \cite{he2021transreid} & ICCV & 2021 & 22.0 & 41.7 & 51.0 & 12.4  \\
 ALBEF \cite{li2021align} & NeurIPS & 2021 & 22.8 & 49.2 & 52.3 & 15.6  \\
 IRRA \cite{jiang2023cross} & CVPR & 2023 & 47.3 & 75.6 & 85.3 & 22.8  \\
 UP-Person \cite{liu2025up} & TCSVT & 2025 & 50.7 & 80.6 & 88.7 & 22.6 \\
 FMFA \cite{yin2025cross} & ACM TOMM & 2025 & 47.8 & 78.9 & 87.6 & 22.1 \\
 GA-DMS \cite{zheng2025gradient} & EMNLP & 2025 & 52.2 & 81.4 & \textcolor{blue}{89.9} & 23.5 \\
 VFE-TPS \cite{shen2025enhancing} & KBS & 2025 & \textcolor{blue}{53.7} & \textcolor{blue}{82.0} & 89.7 & 24.4 \\
 MARS \cite{ergasti2025mars} & ACM TOMM & 2025 & 52.8 & 78.9 & 86.3 & \textcolor{blue}{25.2} \\
PFCVR (Ours) & - & - & \textcolor{red}{55.2} & \textcolor{red}{82.8} &\textcolor{red}{90.3} & \textcolor{red}{26.2}  \\
\hline \toprule [0.5 pt] 
\end{tabular} 
}
\end{table}

\subsection{Comparison on Public Benchmark Datasets} \label{Comparison}
We compare the proposed method with a broad set of baselines spanning two categories: (i) cross-modal retrieval methods published before 2024, including MCANet \cite{ding2024text} (IEEE T-ITS 2024), IRRA \cite{jiang2023cross} (CVPR 2023), ALBEF \cite{li2021align}, and several earlier approaches; (ii) five recent text-to-image person re-identification methods from 2025 that represent the current state of the art on pedestrian benchmarks, namely VFE-TPS \cite{shen2025enhancing}, UP-Person \cite{liu2025up}, FMFA \cite{yin2025cross}, GA-DMS \cite{zheng2025gradient}, and MARS \cite{ergasti2025mars}. All five 2025 methods are retrained on the vehicle datasets under the same evaluation protocol for fair comparison. IRRA serves as the baseline for our implementation. On the T2I-VeRW dataset, certain methods (e.g., MCANet, TFAF, HAT) could not be evaluated due to their dependency on dataset-specific components (such as the three-view masks in MCANet) or the unavailability of source code.

\textbf{Results on T2I-VeRI.} As shown in Table~\ref{quantitativeresults}, PFCVR achieves the best performance across all four metrics. Among the five 2025 baselines, MARS obtains the highest Rank-1 (25.5\%) and mAP (24.9\%); among all competing methods, IRRA holds the highest Rank-5 (57.8\%) and Rank-10 (72.4\%). PFCVR surpasses MARS by 3.7 points on Rank-1 and outperforms IRRA by 2.3 and 3.0 points on Rank-5 and Rank-10, respectively. Notably, none of the five 2025 person retrieval methods surpass the earlier IRRA or MCANet across all four metrics simultaneously when transferred to the vehicle domain. This gap underscores that pedestrian-oriented alignment strategies are suboptimal for vehicles whose distinguishing cues reside in rigid, view-dependent parts rather than body poses or clothing attributes.

\textbf{Results on T2I-VeRW.} Table~\ref{quantitativeresults_new} reports results on our larger-scale benchmark. PFCVR reaches 55.2\% Rank-1, 82.8\% Rank-5, 90.3\% Rank-10, and 26.2\% mAP, leading the second-best method VFE-TPS by 1.5 points on Rank-1 and 0.8 points on Rank-5. On this dataset the 2025 baselines perform more competitively than on T2I-VeRI, likely because the richer annotation granularity and larger training scale of T2I-VeRW partially compensate for the absence of vehicle-specific priors. Nonetheless, PFCVR retains the top rank across all metrics on both benchmarks, indicating that explicit vehicle part alignment and bidirectional mask recovery each address a different source of error and their combination generalizes across dataset scales.

\subsection{Component Analysis} \label{Comparandablation}

Table~\ref{tab:ablation} summarizes the ablation results on T2I-VeRI using the original IRRA model as the baseline. Adding PLFA alone improves Rank-1, Rank-5, Rank-10, and mAP by 1.5, 0.4, 0.2, and 0.6 percentage points, respectively. Further adding BMRIA improves Rank-1, Rank-10, and mAP by an additional 1.5, 1.6, and 0.4 points, though Rank-5 drops slightly by 1.0 points, likely due to the limited training scale where each identity has only 3.2 images on average.

To alleviate this issue, we apply additional data augmentation strategies, including Gaussian noise injection and image brightness adjustment, which expand the training set to four times its original size. As shown in Table~\ref{tab:ablation}, the augmented training data significantly improves the model’s generalization ability and leads to better performance on the test set. Compared with the model trained without augmentation, the full model achieves further improvements of 1.1, 2.9, 1.2, and 0.6 percentage points in Rank-1, Rank-5, Rank-10, and mAP, respectively.

\begin{table}
\center
\small 
\caption{Ablation experiments on the effectiveness of modules. "PLFA" stands for Part-level Local Fine-grained Alignment Module.
"BMRIA" stands for Bidirectional Mask Recovery Implicit Alignment Module.
} 
\label{tab:ablation}
\setlength{\tabcolsep}{2.5pt}{
\begin{tabular}{ccc|cccc}
\hline \toprule [0.5 pt] 
\multicolumn{1}{c}{\multirow{2}{*}{\textbf{\makecell[c]{PLFA}}}} &\multicolumn{1}{c}{\multirow{2}{*}{\textbf{\makecell[c]{BMRIA}}}} &\multicolumn{1}{c|}{\multirow{2}{*}{\textbf{Data Aug.}}} & \multicolumn{4}{c}{\textbf{T2I-VeRI \cite{ding2024text}}}  \\ \cline{4-7} 
\multicolumn{1}{c}{} &
\multicolumn{1}{c}{} &
  \multicolumn{1}{c|}{} &
  \multicolumn{1}{c}{Rank-1} & 
  \multicolumn{1}{c}{Rank-5} &
  \multicolumn{1}{c}{Rank-10}&
  \multicolumn{1}{c}{mAP} \\ \hline
  &   &  & 25.1 & 57.8 & 72.4 & 23.7  \\
 \checkmark &  &  & 26.6 & 58.2 & 72.6 & 24.3 \\
  \checkmark & \checkmark &   & 28.1 & 57.2 & 74.2 & 24.7 \\
    \checkmark & \checkmark &  \checkmark & 29.2 & 60.1 & 75.4 & 25.3 \\
\hline \toprule [0.5 pt] 
\end{tabular} 
}
\end{table}

\subsection{Ablation Study}\label{sec:ablation_study}

\noindent$\bullet$\;\textbf{Comparison of Local Alignment Strategies.}
We compare two designs for incorporating part-level supervision.
\emph{Part MLM} selectively masks part-related tokens in the text and
reconstructs them from image features, driving implicit local alignment
as a byproduct of the self-supervised task.  \emph{PLFA} instead
introduces learnable part-query tokens that first aggregate context from
the full sentence before being aligned with visual part features,
explicitly bridging the gap between isolated part-word semantics and the
global description.

As shown in Table~\ref{localalign}, PLFA improves over Part MLM by 2.9
points on Rank-1, with Rank-5 (+0.7), Rank-10
(+1.1), and mAP (+0.8) also improving.  The gap shows that grounding part queries
in the global sentence context is the decisive factor, not merely the
presence of part-level supervision signals.

\noindent$\bullet$\;\textbf{Image Mask Ratio in BMRIA.}
Table~\ref{tab:maskratio} shows results across ten mask ratios for the
visual reconstruction branch.  Rank-1 peaks at 25\% and generally degrades beyond this
point, though a partial recovery appears between 40\% and 50\% (Rank-1 rises from 25.4 to 26.5) before dropping sharply at 55\%.
Two factors explain this sensitivity: with only 3.2 pairs per
identity on average, the training data cannot support accurate recovery
at high corruption levels; moreover, vehicle parts vary in
spatial footprint, so a single global ratio simultaneously
over-masks small components and under-masks large ones.  A 25\%
ratio best balances task difficulty and data capacity.

\noindent$\bullet$\;\textbf{ITC Loss Weight in PLFA.}
Table~\ref{plfalossweight} shows Rank-1 sensitivity to the ITC loss
weight $\gamma$.  The model achieves the best Rank-1 accuracy at $\gamma{=}0.2$,
which we select as the default since Rank-1 is the primary metric for Re-ID
evaluation.  At this setting, the part-level alignment acts as a useful
regularizer without overriding the global SDM and ID objectives.  Setting $\gamma \geq 0.5$ causes a
Rank-1 drop ranging from 1.5 to 3.7 points, indicating that an
excessively strong local alignment signal disrupts global feature
coherence.  The pronounced sensitivity across all settings is likely
amplified by training-set variance at this data scale.  

\begin{table}
\center
\small 
\caption{Comparison of experimental results of local alignment methods.} 
\label{localalign}
\begin{tabular}{c|cccc}
\hline \toprule [0.5 pt] 
\multicolumn{1}{c|}{\multirow{2}{*}{\textbf{\makecell[c]{Method}}}}  & \multicolumn{4}{c}{\textbf{T2I-VeRI \cite{ding2024text}}}  \\ \cline{2-5}

  \multicolumn{1}{c|}{} &
  \multicolumn{1}{c}{Rank-1} & 
  \multicolumn{1}{c}{Rank-5} &
  \multicolumn{1}{c}{Rank-10}&
  \multicolumn{1}{c}{mAP} \\ \hline
Part MLM & 26.3 & 59.4 & 74.3 & 24.5  \\
PLFA (Ours)  & 29.2 & 60.1 & 75.4  & 25.3 \\
\hline \toprule [0.5 pt] 
\end{tabular} 
\end{table}

\begin{table}
\center
\small 
\caption{Experimental results of visual branch mask ratio in BMRIA. Best results in red, second best in blue.
} 
\label{tab:maskratio}
\begin{tabular}{c|cccc}
\hline \toprule [0.5 pt] 
\multicolumn{1}{c|}{\multirow{2}{*}{\textbf{\makecell[c]{Mask Ratio(\%)}}}}  & \multicolumn{4}{c}{\textbf{T2I-VeRI \cite{ding2024text}}}  \\ \cline{2-5}

  \multicolumn{1}{c|}{} &
  \multicolumn{1}{c}{Rank-1} & 
  \multicolumn{1}{c}{Rank-5} &
  \multicolumn{1}{c}{Rank-10}&
  \multicolumn{1}{c}{mAP} \\ \hline
10 & 26.9 & 56.4 & 74.3 & 24.8  \\
 15  & 28.0 & 56.5 & 74.2  & 24.6  \\
 20  & \textcolor{blue}{28.1} & \textcolor{blue}{59.3} &74.2  & 24.6 \\
 25  & \textcolor{red}{29.2} & \textcolor{red}{60.1} &\textcolor{blue}{75.4}  & \textcolor{blue}{25.3}   \\
 30& 27.2 & 59.0 &73.1  & \textcolor{red}{25.4}  \\
 35 & 26.1 & 58.8 &\textcolor{red}{75.7}  & 25.1  \\
 40 & 25.4 & 57.2 &74.1  & 24.1  \\
 45 & 26.3 & 58.2 &75.3  & 24.7 \\
 50 & 26.5 & 57.2 & 74.8  & 24.9  \\
55 & 25.0 & 55.1 & 70.6  & 23.6  \\
\hline \toprule [0.5 pt] 
\end{tabular} 
\end{table}

\begin{table}
\center
\small 
\caption{Experimental results of ITC loss weight in Part-level Local Fine-grained Alignment Module. Best results in red, second best in blue.
} 
\label{plfalossweight}
\begin{tabular}{c|cccc}
\hline \toprule [0.5 pt] 
\multicolumn{1}{c|}{\multirow{2}{*}{\textbf{\makecell[c]{ITC Loss \\Weight}}}}  & \multicolumn{4}{c}{\textbf{T2I-VeRI \cite{ding2024text}}}  \\ \cline{2-5}

  \multicolumn{1}{c|}{} &
  \multicolumn{1}{c}{Rank-1} & 
  \multicolumn{1}{c}{Rank-5} &
  \multicolumn{1}{c}{Rank-10}&
  \multicolumn{1}{c}{mAP} \\ \hline
0.1 & 26.6 & 58.4 & 75.2 & 24.0  \\
0.2 & \textcolor{red}{29.2} & \textcolor{blue}{60.1} & \textcolor{blue}{75.4} & \textcolor{blue}{25.3}  \\
0.3 & 27.3 & 59.1 & 75.0 & \textcolor{blue}{25.3}  \\
0.4 & 27.6 & 59.1 & 74.3 & \textcolor{red}{25.4}  \\
0.5 & 25.5 & 59.3 & 75.0 & 24.9  \\
0.6 & 26.6 & 59.0 & 74.3 & 25.0  \\
0.7 & 26.9 & 58.0 & 74.1 & 24.4  \\
0.8 & \textcolor{blue}{27.7} & 58.4 & 74.3 & \textcolor{blue}{25.3}  \\
0.9 & 26.3 & 57.6 & \textcolor{blue}{75.4} & 25.0  \\
1.0 & 26.1 & \textcolor{red}{61.6} & \textcolor{red}{76.0} & 24.8  \\
\hline \toprule [0.5 pt] 
\end{tabular} 
\end{table}

\begin{figure}[!htp]
\centering
\includegraphics[width=\linewidth]{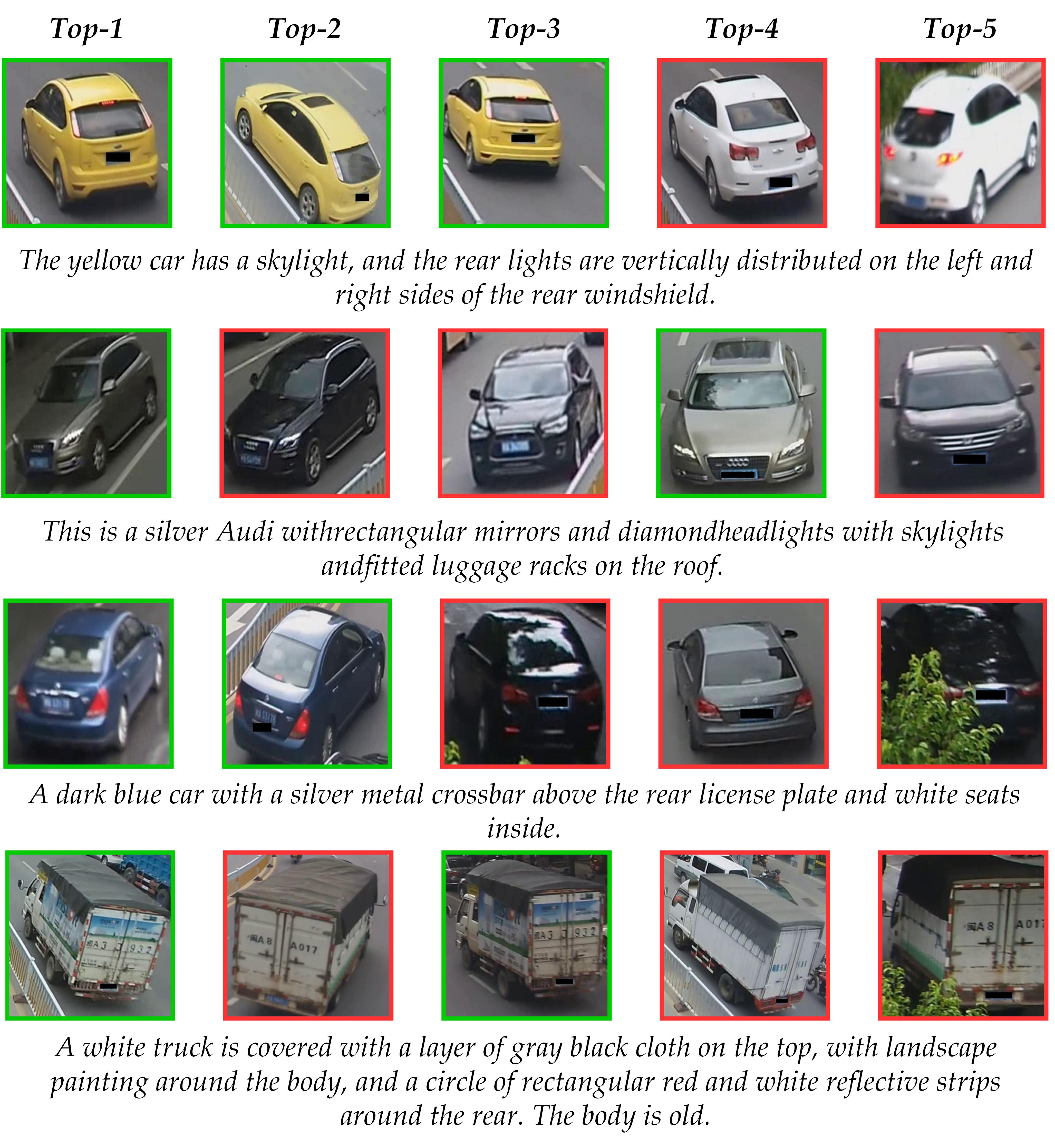}
\caption{Visualization of top-5 retrieval results produced by PFCVR. Green borders indicate correct matches and red borders indicate incorrect matches.} 
\label{fig:visualtop5}
\end{figure}

\subsection{Visualization}\label{sec:visualization} 
Fig.~\ref{fig:visualtop5} shows representative top-5 retrieval results on the T2I-VeRI test set. In the first example, all ground-truth matches appear within the top 3, confirming that PFCVR effectively leverages both holistic appearance and part-level cues. The remaining three examples reveal a recurring error mode: the retrieved vehicles agree with the query on body color and vehicle type, yet deviate in component-level details such as mirror profile or headlight geometry. This gap suggests that further gains are most likely to come from strengthening part-level feature resolution.

\subsection{Limitation Analysis}\label{sec:limitation_analysis}

Although PFCVR achieves competitive results across all benchmarks, the qualitative analysis in Fig.~\ref{fig:visualtop5} exposes a recurring failure mode: vehicles that share the same body shape and color but differ only in texture-level details (grille pattern, badge placement) are often confused. The root cause is that Grounding DINO bounding boxes mark coarse regions without distinguishing the fine textures within them, and the fixed CLIP resolution of $384\times384$ further compresses these small cues. A secondary weakness appears on visually homogeneous categories such as city buses, where appearance cues alone are inherently limited and auxiliary signals (e.g., license-plate regions) would be needed to resolve ambiguity.

\section{Conclusion}

This paper presented PFCVR, a text-to-image vehicle retrieval framework
built on the observation that part-level cues are most useful when they
retain their sentential context.  PLFA materializes this idea through
learnable part-query tokens that fuse part-specific and global text
signals before being aligned with visual regions; BMRIA supplements it
with bidirectional masked reconstruction so that each modality must
attend to the other's fine structure to recover its own masked content.
Together, the two modules achieve 29.2\% Rank-1 on T2I-VeRI (+3.7
over the next best method) and 55.2\% on T2I-VeRW (+1.5), outperforming
both earlier cross-modal baselines and five 2025 person-retrieval
methods retrained on the vehicle domain.  The consistent margin across
two benchmarks of very different scale suggests that the gains stem from
vehicle-specific part modeling rather than dataset-dependent tuning.
Going forward, replacing Grounding DINO bounding boxes with
pixel-level keypoints or texture descriptors should help disambiguate
vehicles that differ only in sub-box details, which currently account
for most failure cases.  Higher-resolution backbones would further
benefit small but informative regions such as rear-window stickers.

\section*{Acknowledgment} 
This work was supported by the National Natural Science Foundation of China under Grant 62102205 and the Anhui Provincial Natural Science Foundation Outstanding Youth Project under Grant 2408085Y032. The authors acknowledge the High-performance Computing Platform of Anhui University for providing computing resources.

\small{ 
\bibliographystyle{IEEEtran}
\bibliography{reference}
}

\end{document}